\documentclass{article}
\usepackage{spconf,amsmath,graphicx}
\usepackage{amsfonts}
\usepackage[ruled,vlined]{algorithm2e}
\usepackage{booktabs}
\usepackage[flushleft]{threeparttable}
\SetKwComment{Comment}{$\triangleright$\ }{}

\title{Mitigating Unintended Memorization in Language Models via Alternating Teaching}

\name{Zhe Liu, Xuedong Zhang, Fuchun Peng}
\address{Meta AI, Menlo Park, CA, USA}

\begin{document}
\ninept
\maketitle

\begin{abstract}
Recent research has shown that language models have a tendency to memorize rare or unique sequences in the training corpora which can thus leak sensitive attributes of user data. We employ a teacher-student framework and propose a novel approach called alternating teaching to mitigate unintended memorization in sequential modeling. In our method, multiple teachers are trained on disjoint training sets whose privacy one wishes to protect, and teachers' predictions supervise the training of a student model in an alternating manner at each time step. Experiments on LibriSpeech datasets show that the proposed method achieves superior privacy-preserving results than other counterparts. In comparison with no prevention for unintended memorization, the overall utility loss is small when training records are sufficient.
\end{abstract}


\section{Introduction}
\label{introduction}
Neural language models (LMs) play important roles in many natural language processing tasks including next word prediction, machine translation, and automatic speech recognition (ASR) \cite{mikolov2010recurrent, chen2015improving, liu2014efficient, kannan2018analysis, irie2019language}. They typically outperform traditional $n$-gram LMs with better capability of modeling long-range dependency.

State-of-the-art LMs typically involve training over large and diverse corpora which might contain sensitive user information, such as addresses and credit card numbers. Recent research has showed that such sensitive information in training datasets can be detected and extracted in unexpected ways \cite{fredrikson2015model, song2019auditing, carlini2019secret, carlini2021extracting, huang2022detecting}. Particularly, LMs are prone to \emph{unintentionally memorize} rare or unique sequences of data, and when being prompted appropriately, they will be able to emit the memorized text verbatim \cite{carlini2022quantifying}. This is undesirable because such memorization violates privacy by exposing user information. Therefore, providing privacy guarantees to LM training has become a critical problem and it calls for advanced mitigation techniques for unintended memorization in LMs.

In this paper, we employ a teacher-student framework and propose a novel method called \emph{alternating teaching} to mitigate the issue of unintended memorization in sequential modeling. In our approach, multiple teachers are trained on disjoint training data (e.g. data from different users) whose privacy one wishes to protect , and teachers' predictions are utilized as soft labels to supervise the training of a student model. Unlike teacher ensemble and aggregation methods, at each word-level time step of student model training, we only choose one teacher to provide supervision. That is, teachers are selected in an alternating manner through randomization or permutation. Finally, only the student model is published while all teachers are kept private.

Intuitively, in most scenarios any piece of sensitive information is only contained in the training text of one specific user, and is thus exposed to one teacher model. Then alternating teacher selection at each time step breaks the semantic and linguistic connections between consecutive words in any private sequences, but is still able to learn from common and non-sensitive word combinations that are exposed to all teacher models. Thus, this technique can reduce the level of memorization without appreciable loss of overall utility in the model.

Our approach is inspired by text generation models which generate words in a sequence step-by-step and left-to-right. One promising method for addressing memorization via text generation is to use one teacher to generate a word at each time step and the word with its historical contexts acts as the next input for another teacher. However, this strategy is costly in computation especially when hundreds of millions of sentences need to be generated. Under the knowledge distillation framework, the proposed alternating teaching approach is more efficient and scalable.

We make the following contributions: (1) introducing the new alternating teaching based teacher-student framework for effective mitigation of unintended memorization in LMs; (2) studying the effect of various knowledge distillation mechanisms on alleviating memorization, which includes the use of public training corpora and adding random noises to teachers' output distributions; and (3) providing empirical results and analyses on comparing the utility and privacy protection of various teacher-student learning based approaches.

The rest of the paper is organized as follows. We review related work in Section~\ref{relatedwork}. Section \ref{methodology} describes the details of our proposed alternating teaching method. Next, Section \ref{experiments} shows the experiments and results for LM and ASR tasks on the \emph{LibriSpeech} data \cite{panayotov2015librispeech}. We conclude in Section \ref{conclusion}.

\section{Related Work}
\label{relatedwork}
Privacy protection is becoming crucial in machine learning research. One direction in this area is \emph{private aggregation of teacher ensembles} (PATE) \cite{papernot2016semi,papernot2018scalable}, which transfers to a student model the knowledge of an ensemble of teacher models, with strong privacy guaranteed by noisy aggregation and vote counts of teachers' answers. More recently, authors in \cite{tian2022seqpate} adapts PATE to text generation tasks while satisfying \emph{differential privacy} (DP) \cite{dwork2006calibrating, dwork2014algorithmic}. Our work differs from PATE and its variants in alternating teacher selection instead of aggregation mechanisms. Moreover, we focus on the empirical measurement of unintended memorization rather than DP-based privacy analysis. Another line of research on privacy-preserving methods is federated learning (FL) \cite{konevcny2016federated2,mcmahan2017communication,ramaswamy2020training,thakkar2021understanding}. Apart from these works, authors in \cite{feldman2020does} explores how memorization relates to generalization in learning.

\section{Methodology}
\label{methodology}
\subsection{The Alternating Teaching Framework}
To help mitigate unintentionally memorization in LMs, the proposed alternating teaching framework transfers knowledge from alternating teacher models trained on partitions of the data to a student model. Our method consists of three key parts: (1) multiple teacher models, (2) teacher selection mechanism, and (3) a student model.

\subsubsection{Teacher Models}
Each teacher model is an LM trained independently on a subset of the data. The data is partitioned into disjoint subsets by users to ensure no pair of teachers will have trained on data from the same user. In other words, all records from any user is only included in the training corpus of one specific teacher model.

More formally, let $\mathcal{D}_k=\{\mathcal{D}_k^{\text{pri}}\cup\mathcal{D}_k^{\text{pub}}\}$ be the training corpus from the $k$th user, where $\mathcal{D}_k^{\text{pri}}$ is a set of records with sensitive information and $\mathcal{D}_k^{\text{pub}}$ is a corpus without sensitive information. Then let $\mathcal{D}=\cup_{k=1}^K \mathcal{D}_k$ be the entire training set over all users' data. Without the loss of generality, assuming the training set is partitioned into $M$ disjoint subsets by users, denote $\mathcal{B}_{m}=\cup_{k=(m-1)d+1}^{md} \mathcal{D}_k$ for each $m=1,\ldots, M$ and $d=[K/M]$. Thus we have $\mathcal{D}=\cup_{m=1}^M \mathcal{B}_m$ and $\mathcal{B}_{m'}\cap\mathcal{B}_{m''}=\emptyset$ for any $m'\neq m''$.

The $m$th teacher model, denoted as $f_m(\theta_m)$ with $\theta_m$ being the weights, is trained using text set of $\mathcal{B}_m$. Cross entropy (CE) loss is usually used for LM training. Given any training example with $T$ words, $(w_1,w_2,\ldots,w_T) \in \mathcal{B}_m$, the following shows this function at step $t$
\begin{align}
\mathcal{L}^{\text{CE}}_t(\theta_m)=-\sum_{w\in V} \textbf{1}\{w=w_t\}\cdot\log p_{\theta_m}(w|w_{1:t-1})
\end{align}
where $V$ is the vocabulary set and $f_m(\theta_m)$ predicts a word $w$ with a probability $p_{\theta_m}(w|w_{1:t-1})$ at step $t$.

Thus we obtain a set of teacher models $\{f_m(\theta_m)\}_{m=1}^M$. Note that any teacher is not privacy-preserving and susceptible to unintended memorization since it trains on combined sets of sensitive corpus and non-sensitive corpus.

\subsubsection{Alternative Teaching Mechanism}
Once all teacher LMs are trained, given any $(w_1,w_2,\ldots,w_T) \in \mathcal{D}$, each teacher $\{f_m(\theta_m)\}_{m=1}^M$ conducts the inference on it and outputs probability distribution $p_{\theta_m}(
\cdot|w_{1:t-1})$ over all words in the vocabulary at step $t$. Here, we discuss how these predictions can be combined to provide supervisions.

One promising approach is the aggregation mechanism where at each step, the predicted probabilities from teachers are averaged on each word in the vocabulary. In particular, the following ensemble teacher output is used to supervise a student model at step $t$
\begin{align}
g^{\text{agg}}(\cdot|w_{1:t-1})=\frac{1}{M}\sum_{m=1}^M p_{\theta_m}(\cdot|w_{1:t-1})
\end{align}
Then a student model is trained on this aggregated output of the $M$ teachers, such that it learns to accurately mimic the ensemble. Intuitively, this aggregation strategy ensures no single teacher and thus no single user's dataset dictates the student's training. This will help alleviate any unintended memorization. However, one disadvantage of this approach is that when the presence of some private sequence in one specific teacher is very strong and even dominating, simply taking the average over the probabilities of all teachers might not be adequate to provide a full coverage and still reveal such sensitive information to the student.

In the newly proposed alternative teaching mechanism, at each time step we only leverage the prediction output from one teacher rather than using all teachers' aggregation, and alternate the choices of teachers over different steps. This can be performed through randomization or fixed permutation.

In the randomization based teacher selection, for each step $t$, we randomly generate $r^{\text{random}}(t)\in\{1,2,\ldots,M\}$ and the corresponding teacher model is chosen as the supervisor. It aims to disconnect consecutive words in private sequences but generally has no issues in learning common and non-sensitive sequences that are present in majority of teachers.

The randomized teacher selection happens in each step at every batch during training. A more restricted teacher selection strategy is through permutation but kept the chosen order fixed over the entire training process. Specifically, let $\pi(M)$ be a random permutation of the sequence $\{1,2,\ldots,M\}$, then the teacher index at step $t$, denoted as $r^{\text{perm}}(t)$, is chosen as the $j$th element of $\pi(M)$, where $j=(t\bmod M)$ if the corresponding remainder is non-zero; otherwise $j=M$. The assignment of $r^{\text{perm}}(t)$ stays intact across different batches and epochs.

In either case, let $r(t)$ be the selected teacher index at step $t$, then we write
\begin{align}
g^{\text{alt}}(\cdot|w_{1:t-1})=p_{\theta_{r(t)}}(\cdot|w_{1:t-1})
\end{align}
as the predicted distribution which is used to supervise the student model at step $t$.

\subsubsection{Student Model}
Since any training corpora are naturally labeled for LM task, the student model, denoted by $f(\theta)$, is supervised by both the labels from its training set and combined teachers' outputs. Then for any sequence $(w_1,w_2,\ldots,w_T)$, the following computes the loss function consisting of two parts
\begin{align}
\label{loss}
\mathcal{L}(\theta)&=\sum_{t=1}^T\left((1-\lambda)\cdot \mathcal{L}_t^{\text{CE}}(\theta)+\lambda\cdot \mathcal{L}_t^{\text{KL}}(\theta)\right) \\
\mathcal{L}^{\text{CE}}_t(\theta):&=-\sum_{w\in V} \textbf{1}\{w=w_t\}\cdot\log p_{\theta}(w|w_{1:t-1}) \\
\mathcal{L}^{\text{KL}}_t(\theta):&=D_\text{KL}(g^{\text{alt}}(\cdot|w_{1:t-1})\,||\,p_{\theta}(\cdot|w_{1:t-1})) \label{loss:kl}
\end{align}
where $D_\text{KL}(P\,||\,Q)$ represents the Kullback–Leibler divergence between distributions $P$ and $Q$, and $\lambda$ is a hyperparameter which balances the two parts of $\mathcal{L}_t^{\text{CE}}(\theta)$ and $\mathcal{L}_t^{\text{KL}}(\theta)$.

The student model can be trained on any auxiliary, non-sensitive corpora, including publicly available collections of text data. However, when such dataset is not available or the student model suffers from utility loss due to distillation, the original set $\mathcal{D}$ can still be used to train the student model. In that case, the hyperparameter $\lambda$ shall be set as 0 since $\mathcal{D}$ contains private information and we do not want it is directly exposed to the student model. In that case, the student model fully learns from combined teachers' outputs.

\subsection{The Gaussian Noise Mechanism}
Building upon the alternating teaching framework described above, random noises can be added to the outputs from teacher models so that they can further mask the presence of private sequences and thus make sensitive information less susceptible to leakage. We apply the Gaussian mechanism which adds noise independently sampled from a Gaussian distribution $\mathcal{N}(0,\sigma^2)$ to each coordinate of the predicted probabilities from teachers, after which the re-normalization over vocabulary space is needed. The hyperparameter $\sigma$ governs the strength of privacy protection. Specifically, the teacher supervision part $\mathcal{L}^{\text{KL}}_t(\theta)$ in~(\ref{loss:kl}) can be adjusted as
\begin{align}
D_\text{KL}(s(g^{\text{alt}}(\cdot|w_{1:t-1})+\mathcal{N}(0,\sigma^2))\,||\,p_{\theta}(\cdot|w_{1:t-1}))
\end{align}
where $s(\cdot)$ is a normalization function over the vocabulary such that all probabilities with added noises are truncated to non-negative and their sum equals to 1 after normalization.

\section{Experiments}
\label{experiments}
\subsection{Datasets}
Our experiments use the LibriSpeech data \cite{panayotov2015librispeech} and its extended text-only corpus \cite{openSLR11}:
\begin{itemize}
  \item LibriSpeech ASR corpus and text transcripts. It is a corpus of around 1000 hours of 16kHz read English audiobooks. The dataset consists of {train}, {validation}, and {test} splits, which contain 281K, 6K, and 6K utterances from approximately 2400 speakers, respectively;
  \item LibriSpeech extended text-only corpus. It is from 14500 public domain books which contains around 40M sentences. The dataset is only for LM training purpose.
\end{itemize}

In the next subsection, we will describe how these datasets are augmented with ``private sequences'' such that we can measure the performance of mitigating unintended memorization over different methods.

In some portion of our experiments, we also utilize the training dataset from Wikitext-103 \cite{merity2016pointer}. This is treated as an auxiliary and non-sensitive public corpus, on which the student LM is trained.

\subsection{Canaries}
To measure the level of unintended memorization in LMs, we build on the “secret sharer" framework introduced in \cite{carlini2019secret}. Specifically, random textual sequences, called \emph{canaries}, are inserted into a training corpus, and a model trained on this corpus is then analyzed to measure the frequencies of having these canaries memorized. Here, the canaries aim to mimic sensitive data.

The procedure of inserting canaries into LibriSpeech datasets is described as follows:
\begin{enumerate}
  \item[(1)] First, each record is assigned a user ID. For the LibriSpeech ASR corpus, user ID of any utterance is just the speaker id. For the LibriSpeech text-only corpus, we randomly shuffle all the records, and create synthetic users where each user owns 100 records, assigned sequentially from the shuffled set;
  \item[(2)] Next, we randomly pick 100 users for each of the two LibriSpeech datasets. For each user, a random 5-word canary is generated which simulates the ``private sequence'' from that specific user. No canary is shared by different users. Note that each word in any generated canary is among the vocabulary set of LMs;
  \item[(3)] For each generated canary, we insert it into the LibriSpeech training corpora at a certain frequency (i.e. number of times it is repeated). Specifically, the 100 canaries (from 100 users) are evenly partitioned into 4 groups with canaries' repeating frequencies being 5\%, 10\%, 50\%, and 100\%, respectively. For each canary, let $n_u$ be the number of training records in the corresponding user and $p_u$ be the repeating frequency based on the group it belongs to, then $p_u \cdot n_u$ is the number of records that the canary is inserted into the training corpora of the corresponding user.
\end{enumerate}
The procedure illustrated above is intended to simulate real-world scenarios where any occurrences of user-specific unique or rare out-of-distribution canaries are typically limited to a very small fraction of users, but these users can exhibit either low or high usage of those canaries \cite{thakkar2021understanding}.

Given a prefix of a canary, we use the following two techniques to evaluate the mitigation of unintended memorization for any LM:
\begin{itemize}
  \item Beam Search (BS). We leverage a greedy beam search to see if the canary is included in the top 100 most-likely 5-word continuations from the 1-word prefix of the canary;
  \item Random Sampling (RS). We say any canary is unintentionally memorized by a LM if the canary has the least perplexity among 1000 random suffixes, given the 2-word prefix of the canary.
\end{itemize}
In our experiments, we report the frequencies of times that the 100 generated canaries are detected by BS or RS in any LM.

\subsection{Setups}
With the generated canaries inserted into the two LibriSpeech training sets, LMs are trained on the text corpora with their perplexity (PPL) measured on the test split of LibriSpeech ASR corpus. The level of unintended memorization is evaluated using the BS and RS techniques described above.

The LM in our experiments is LSTM based with embeddings dimension 300, and 2 layers of 1500 hidden units. The word vocabulary set is around 10K. We use Adam optimizer and early stopping based on the validation set of LibriSpeech ASR corpus.

In our experiments, we consider the following approaches in the comparison of utility and unintended memorization mitigation:
\begin{itemize}
  \item The \texttt{Baseline} LM is directly trained on LibriSpeech data (either ASR corpus or text-only corpus) with canaries;
  \item \texttt{Baseline(1T)} refers to the student LM with knowledge distilled from a single teacher model. Here, both teacher and student models are trained on LibriSpeech data with canaries;
  \item \texttt{Agg} represents the student model supervised by aggregation based teacher ensembles. We use the notations of \texttt{Agg(2T)} and \texttt{Agg(5T)} to denote there are 2 and 5 teachers for knowledge distillation, respectively. Again, all teachers and the student are trained on LibriSpeech corpora with canaries;
  \item \texttt{Alt-Random} and \texttt{Alt-Perm} are our proposed approaches of alternating teaching with randomization and fixed permutation strategies for selecting teachers per step, respectively.
\end{itemize}
In each of knowledge distillation based methods, the parameter of $\lambda$ is set to 0. In other words, the student model is only supervised by  teachers' outputs.

On the test split of LibriSpeech ASR corpus, we also evaluate the ASR performance, in terms of word-error-rate (WER), with the LMs being used as second-pass rescorers on the generated 20-best hypotheses. The ASR model is a RNN-T model with the Emformer encoder \cite{emformer2021streaming}, LSTM predictor, and a joiner. It has around 80 million parameters and is trained from scratch using the train split of LibriSpeech ASR corpus. Note that we only measures the impact on WER when LMs are trained on LibriSpeech text-only corpus since they will be more effective in rescoring. This is because the ASR model does not include such text-only corpus in its model training.

\subsection{Results}
We first measure the performance of (student) LMs trained on the LibriSpeech ASR corpus with canaries. All teacher models are also trained on it. Table~\ref{tab:lib} shows the PPL results on the test split as well as the percentage of canaries being uncovered by BS and RS techniques. Here, the 100 canaries are partitioned into two categories with low repeating frequencies and high repeating frequencies, and we report their results separately. From the results
\begin{itemize}
  \item In the \texttt{Baseline} methods, canaries are substantially memorized by the LMs. Memorization is detectable even for canaries that appear only a few times in the training corpus;
  \item It is expected to see that having more teacher models leads to stronger mitigation of memorized canaries in all methods. Particularly, the proposed \texttt{Alt} performs better than \texttt{Agg}, and \texttt{Alt-Perm} has the fewest canaries detected;
  \item Degradation on PPL are observed when student models are supervised by multiple teachers. This is expected since LibriSpeech ASR corpus only contains less than 300K training records, thus partitioning them into multiple disjoint sets for training teacher models will cause accuracy loss due to insufficiency of training data.
\end{itemize}

\begin{table}[ht!]
  \vspace{-0.4cm}
  \caption{Results for LMs trained on LibriSpeech ASR corpus. PPL and percentages of canaries detected by BS and RS are reported. Low: group of canaries with low (5\% or 10\%) repeating frequencies; High: canaries with high (50\% or 100\%) repeating frequencies.}
  \centering
  \resizebox{0.9\columnwidth}{!}{%
  \begin{threeparttable}
  \begin{tabular}{l|r|rr|rr}
    \toprule
    & \emph{Utility}
    & \multicolumn{2}{|c}{\emph{BS}} 
    & \multicolumn{2}{|c}{\emph{RS}}\\
    \cmidrule(r){2-6}
    \emph{Method} & \emph{PPL} & \emph{Low} & \emph{High} & \emph{Low} & \emph{High} \\
    \midrule
    \texttt{Baseline} & 76.4 & 95\% & 100\% & 100\% & 100\% \\
    \midrule    
    \texttt{Baseline(1T)} & 76.0 & 92\% & 100\% & 100\% & 100\% \\
    \midrule
    \texttt{Agg(2T)} & 83.8 & 66\% & 100\% & 100\% & 100\% \\
    \texttt{Alt-Random(2T)} & 84.4 & 62\% & 100\% & 100\% & 100\% \\
    \texttt{Alt-Perm(2T)} & 86.6 & 0\% & 0\% & 100\% & 100\% \\   
    \midrule
    \texttt{Agg(5T)} & 100.5 & 0\% & 32\% & 48\% & 100\% \\
    \texttt{Alt-Random(5T)} & 101.9 & 0\% & 4\% & 2\% & 98\% \\
    \texttt{Alt-Perm(5T)} & 107.0 & 0\% & 0\% & 0\% & 18\% \\  
    \bottomrule
  \end{tabular}
  \end{threeparttable}
  }
  \label{tab:lib}
\end{table}

Table~\ref{tab:wiki} displays the results where all the teachers are trained using LibriSpeech ASR corpus with  canaries, but the student LMs are trained on Wikitext-103 data with teachers' supervision. The observations on the comparison of different methods are similar to the ones in Table~\ref{tab:lib}, where \texttt{Alt-Perm} has the smallest number of canaries being detected. Although the PPL results do not change much from the ones in Table~\ref{tab:lib}, we can see that training the student models using an auxiliary and non-sensitive dataset here  achieves reduced memorization.

\begin{table}[ht!]
  \caption{Results for student LMs trained on Wikitext-103; Teacher LMs are still trained on LibriSpeech ASR corpus.}
  \centering
  \resizebox{\columnwidth}{!}{%
  \begin{threeparttable}
  \begin{tabular}{l|r|rr|rr}
    \toprule
    & \emph{Utility}
    & \multicolumn{2}{|c}{\emph{BS}} 
    & \multicolumn{2}{|c}{\emph{RS}}\\
    \cmidrule(r){2-6}
    \emph{Method} & \emph{PPL} & \emph{Low} & \emph{High} & \emph{Low} & \emph{High} \\
    \midrule
    \texttt{Baseline} & 76.4 & 95\% & 100\% & 100\% & 100\% \\
    \midrule    
    \texttt{Baseline(Wiki,1T)} & 76.5 & 92\% & 100\% & 100\% & 100\% \\
    \midrule
    \texttt{Agg(Wiki,2T)} & 84.6 & 14\% & 48\% & 100\% & 100\% \\
    \texttt{Alt-Random(Wiki,2T)} & 84.9 & 4\% & 30\% & 100\% & 100\% \\
    \texttt{Alt-Perm(Wiki,2T)} & 87.9 & 0\% & 0\% & 94\% & 92\% \\   
    \midrule
    \texttt{Agg(Wiki,5T)} & 101.2 & 0\% & 12\% & 18\% & 98\% \\
    \texttt{Alt-Random(Wiki,5T)} & 101.4 & 0\% & 6\% & 2\% & 88\% \\
    \texttt{Alt-Perm(Wiki,5T)} & 108.5 & 0\% & 0\% & 0\% & 12\% \\  
    \bottomrule
  \end{tabular}
  \end{threeparttable}
  }
  \label{tab:wiki}
  \vspace{-0.4cm}
\end{table}

Next, we measure the performance of (student) LMs trained on LibriSpeech text-only corpus with inserted canaries. The test split of LibriSpeech ASR corpus is used for evaluating the PPL and WER results. Seen from Table~\ref{tab:text}, the utility gaps are relatively small over methods with different numbers of teachers, which can be explained by the large training corpus of 40M sentences. Particularly, WERs only differ in less than 1\% comparing \texttt{Agg} or \texttt{Alt} with  \texttt{Baseline}. Thus, all these approaches tend to match the baseline utility while being empirically less prone to memorization. Again, we notice that \texttt{Alt-Perm(5T)} achieves the strongest mitigation of memorization comparing with others, and \texttt{Alt-Random(5T)} obtains less memorization than \texttt{Agg(5T)}.

\begin{table}[ht!]
  \vspace{-0.4cm}
  \caption{Results for LMs trained on LibriSpeech text-only corpus.}
  \centering
  \resizebox{\columnwidth}{!}{%
  \begin{threeparttable}
  \begin{tabular}{l|rr|rr|rr}
    \toprule
    & \multicolumn{2}{|c}{\emph{Utility}}
    & \multicolumn{2}{|c}{\emph{BS}} 
    & \multicolumn{2}{|c}{\emph{RS}}\\
    \cmidrule(r){2-7}
    \emph{Method} & \emph{PPL} & \emph{WER} & \emph{Low} & \emph{High} & \emph{Low} & \emph{High} \\
    \midrule
    \texttt{NoLM} & -\;\; & 6.92 & -\;\; & -\;\;\; & -\;\;\; & -\;\;\; \\
    \midrule
    \texttt{Baseline} & 46.9 & 6.59 & 2\% & 78\% & 38\% & 100\% \\
    \midrule    
    \texttt{Baseline(1T)} & 48.5 & 6.61 & 2\% & 68\% & 22\% & 98\% \\
    \midrule
    \texttt{Agg(2T)} & 49.2 & 6.61 & 2\% & 64\% & 18\% & 94\% \\
    \texttt{Alt-Random(2T)} & 49.4 & 6.62 & 0\% & 50\% & 12\% & 94\% \\
    \texttt{Alt-Perm(2T)} & 49.4 & 6.62 & 0\% & 0\% & 14\% & 92\% \\   
    \midrule
    \texttt{Agg(5T)} & 50.5 & 6.64 & 0\% & 12\% & 4\% & 72\% \\
    \texttt{Alt-Random(5T)} & 51.0 & 6.64 & 0\% & 0\% & 2\% & 40\% \\
    \texttt{Alt-Perm(5T)} & 51.0 & 6.64 & 0\% & 0\% & 0\% & 24\% \\  
    \bottomrule
  \end{tabular}
  \end{threeparttable}
  }
  \label{tab:text}
\end{table}

Lastly, we study the effect of adding Gaussian noises on top of the \texttt{Alt-Perm} framework. Seen from the results in Table~\ref{tab:noisy}, the \texttt{Alt-Perm(5T)} method with noise scale parameter $\sigma=1e^{-4}$ has no canaries detected by BS or RS, while the WER is compromised by around 1.5\% compared with \texttt{Baseline}.

\begin{table}[ht!]
  \vspace{-0.4cm}
  \caption{Results for LMs trained on LibriSpeech text-only corpus, with Gaussian noise mechanism being applied.}
  \centering
  \resizebox{\columnwidth}{!}{%
  \begin{threeparttable}
  \begin{tabular}{l|rr|rr|rr}
    \toprule
    & \multicolumn{2}{|c}{\emph{Utility}}
    & \multicolumn{2}{|c}{\emph{BS}} 
    & \multicolumn{2}{|c}{\emph{RS}}\\
    \cmidrule(r){2-7}
    \emph{Method} & \emph{PPL} & \emph{WER} & \emph{Low} & \emph{High} & \emph{Low} & \emph{High} \\
    \midrule
    \texttt{Alt-Perm(5T)} & 51.0 & 6.64 & 0\% & 0\% & 0\% & 24\% \\ 
    \midrule
    \texttt{Alt-Perm(5T,$\sigma=1e^{-5}$)} & 53.1 & 6.65 & 0\% & 0\% & 0\% & 10\% \\ 
    \texttt{Alt-Perm(5T,$\sigma=1e^{-4}$)} & 59.2 & 6.69 & 0\% & 0\% & 0\% & 0\% \\ 
    \bottomrule
  \end{tabular}
  \end{threeparttable}
  }
  \label{tab:noisy}
\end{table}

\section{Conclusion}
\label{conclusion}
In this work, we propose the alternating teaching method to mitigate unintended memorization in sequential modeling. With experiments on LibriSpeech datasets, we show this approach achieves stronger mitigation than other counterparts and significantly reduces memorized sequences. Compared with the baselines without protections for memorizing private data, the overall quality of proposed method is not compromised when there exists sufficient training data.

Future work might include extending the proposed framework to user-level DP-based privacy analysis. 


\bibliographystyle{IEEEbib}
\bibliography{refs}

\end{document}